\documentclass[10pt, a4paper]{article}

\usepackage[final]{lrec2026}
\usepackage{float}
\usepackage{booktabs}
\usepackage{multirow}
\usepackage{adjustbox}
\usepackage[table]{xcolor}
\usepackage{colortbl}
\usepackage{float}
\usepackage{caption}
\definecolor{low}{RGB}{230,245,255}
\definecolor{mid}{RGB}{115,179,216}
\definecolor{high}{RGB}{8,48,107}
\newcommand{\heat}[1]{%
    \begingroup
    \pgfmathparse{#1}%
    \ifdim\pgfmathresult pt < 0.2pt \cellcolor{low!30}\else%
    \ifdim\pgfmathresult pt < 0.4pt \cellcolor{mid!30}\else%
    \ifdim\pgfmathresult pt < 0.6pt \cellcolor{mid!60}\else%
    \ifdim\pgfmathresult pt < 0.8pt \cellcolor{high!40}\else%
    \cellcolor{high!70}\fi\fi\fi\fi%
    #1
    \endgroup
}
\usepackage{pgf}
\usepackage{booktabs}
\usepackage{pgfplots}
\pgfplotsset{compat=1.18}
\usepackage{multirow}
\usepackage{wrapfig}
\usepackage{comment}
\usepackage{nicematrix}
\usepackage{tabularx}
\usepackage[margin=2cm]{geometry}
\usepackage{multicol}
\usepackage{tikz}
\usepackage{adjustbox}
\usepackage{array}
\usepackage{float}
\usepackage{tabularx, booktabs}
\usepackage{booktabs,tabularx,makecell,siunitx}
\newcolumntype{Y}{>{\raggedright\arraybackslash}X}
\usepackage{algorithm}
\usepackage[noend]{algpseudocode}
\usepackage{titlesec}
\usepackage{enumitem}
\titlespacing*{\subsection}{0pt}{0.5em plus 0.2em minus 0.2em}{0.3em}
\titlespacing*{\subsubsection}{0pt}{0.4em plus 0.2em minus 0.2em}{0.2em}
\makeatletter
\def\BState{\State\hskip-\ALG@thistlm}
\makeatother
\usetikzlibrary{shapes.geometric, arrows.meta, positioning}

\usepackage{array,multirow,hhline}
\renewcommand{\arraystretch}{1.2}
\title{FIBER: A Multilingual Evaluation Resource for Factual Inference Bias}
\usepackage{array}
\newcolumntype{C}[1]{>{\centering\arraybackslash}m{#1}}

\name{Evren Ayberk Munis$^{1}$, Deniz Yılmaz$^{2}$, Arianna Muti$^{3}$, Çağrı Toraman$^{2}$}

\address{
$^{1}$Politecnico di Torino, Italy\\
$^{2}$Middle East Technical University, Computer Engineering Department, Turkey\\
$^{3}$Bocconi University, Italy\\
$^{1}$\texttt{evrenayberk.munis@studenti.polito.it}\\
$^{2}$\texttt{deniz.yilmaz\_12@metu.edu.tr}, \texttt{ctoraman@metu.edu.tr}\\
$^{3}$\texttt{arianna.muti@unibocconi.it}
}

\usepackage{array}
\abstract{
Large language models are widely used across domains, yet there are concerns about their factual reliability
and biases. Factual knowledge probing offers a systematic means to evaluate these aspects. Most existing
benchmarks focus on single-entity facts and monolingual data. We therefore present FIBER, a multilingual benchmark for
evaluating factual knowledge in single- and multi-entity settings. The dataset includes sentence completion, question-answering, and object-count prediction tasks in English, Italian, and Turkish. Using FIBER, we examine whether the prompt language induces inference bias in entity selection and how large language models perform on multi-entity versus single-entity questions. The results indicate that the language of the prompt can influence the model’s generated output, particularly for entities associated with the country corresponding to that language. However, this effect varies across different topics such that 31\% of the topics exhibit factual inference bias score greater than $0.5$. Moreover, the level of bias differs across languages such that Turkish prompts show higher bias compared to Italian in 83\% of the topics, suggesting a language-dependent pattern. Our findings also show that models face greater difficulty when handling multi-entity questions than the single-entity questions. Model performance differs across both languages and model sizes. The highest mean average precision is achieved in English, while Turkish and Italian lead to noticeably lower scores. Larger models, including Llama-3.1-8B and Qwen-2.5-7B, show consistently better performance than smaller 3B–4B models.
 \\ \newline  }

\begin{document}

\maketitleabstract

\section{Introduction}
Despite the widespread adoption of large language models (LLMs), they often exhibit unreliability and a tendency to generate false or fabricated information, which is known as LLM hallucination \citep{huang2024surveyhallucination}. Consequently, assessing their reliability has become a crucial research objective. In this context, Factual Knowledge Probing plays a fundamental role, serving as a systematic method to evaluate whether models accurately store and retrieve factual information without bias.
\begin{figure}[h!]
  \centering
  \includegraphics[width=0.8\columnwidth]{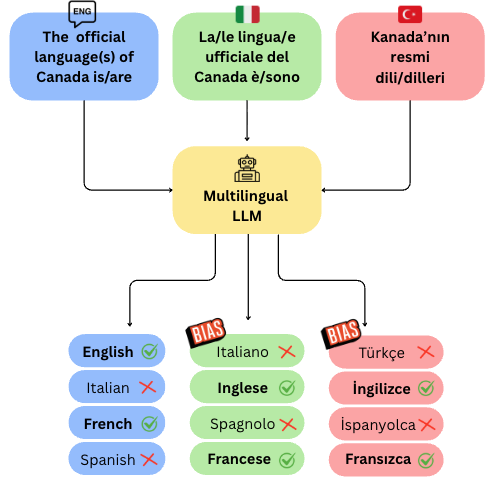} 
  \caption{Multilingual prompts are provided to a multilingual language model, and candidate answers are ranked in descending order according to their cumulative log-probability scores. In the figure, gold answers are marked with a green check, incorrect predictions with a red cross, and inference-biased answers with a bias label indicating cases where the model’s prediction is influenced by the language of the prompt rather than by factual correctness.}
  \label{fig:figure1}
\end{figure}

One of the notable biases observed in LLMs is their tendency to favor the language of the prompt when generating responses. This phenomenon is referred as inference bias ( \citep{kim2025geopolitical,li2024geopolitical}). Inference bias is measured by analyzing how frequently the model generated words or entities related to the region associated with the prompt language, using human or model-based evaluators.
In this study, we extend this concept to the factual knowledge probing setting by introducing a new term, \textbf{Factual Inference Bias}. While inference bias focuses on the frequency of language-aligned responses in generated text, Factual Inference Bias examines the same phenomena by asking factual questions and evaluating by the probability ranking of ground truth tokens. Specifically, it captures cases where the model assigns higher likelihoods to factual entities that are geographically aligned with the prompt language. For instance, while the official languages of Canada are English and French, the model assigns higher probabilities to answers associated with the prompt language when the input is provided in Italian or Turkish as shown in \autoref{fig:figure1}.

\citet{kim2025geopolitical}’s analysis is also limited to question–answering tasks with single-entity answers. However, in real-world scenarios, a single subject can be linked to multiple objects, leading to multi-entity answers (e.g. official languages of countries). Moreover, the degree of bias may vary depending on the prompt type, such as question-answering versus sentence completion. Existing factual knowledge probing benchmarks \citep{petroni2019lama,elazar2021pararel,kwiatkowski2019natural,lin2021truthfulqa} mainly focus on single-entity questions, while multi-entity datasets are rare and often monolingual. For instance, MyridLAMA \citep{zhao2024memorizing} explores various prompt formats; however, it focuses on different structural variations of subjects and answers rather than having subjects that have  one-to-many relationships. Therefore, there is still no multilingual dataset that combines both single- and multi-entity answers across comparable topics, making it difficult to evaluate and compare model behavior under such conditions. To address this gap, we investigate the following research questions:

\paragraph{RQ1} Does the language of the prompt induce Factual Inference Bias in LLMs in different topics, such that models tend to favor entities associated with the country or region where the prompt language is mostly spoken?

\paragraph{RQ2} Given that most existing benchmarks are limited to datasets containing only single-entity answers, how do LLMs perform when faced with multi-entity questions?

Factual Knowledge Probing is an experimental framework designed to evaluate how effectively a model recalls and utilizes real-world factual information when answering queries. To assess factuality in LLMs, researchers commonly compare performance across two task types: open-ended generation \citet{youssef2023givemethefacts}, such as sentence completion or continuation prompts, and knowledge-seeking question-answering (QA) tasks {\citet{youssef2023givemethefacts}}. The performance of a model may vary significantly between these formats and across different languages. Therefore, evaluating both task types and multiple linguistic settings is important for obtaining a comprehensive understanding of a model’s overall factual knowledge capabilities.

We examine these two research questions using a new multilingual dataset called FIBER\footnote{The code and dataset to replicate the experiments can
be found at \url{https://github.com/metunlp/fiber}\label{fn:shared}}, which stands for Factual Inference Bias Evaluation Resource. FIBER includes both question-answering and sentence completion tasks for single- and multi-entity answers in English, Italian, and Turkish. As shown in \autoref{fig:figure1}, multilingual prompts are provided to the model, and the log probabilities of a predefined surface set—containing all possible answers across subjects—are computed. Instead of generating new tokens, we analyze the model’s assigned probabilities to these candidate answers. The ranked list of probabilities is then evaluated using the Average Precision (AP)\citep{zhu2004averageprecision} metric, which measures how well the model ranks the ground-truth answers among all possible candidates. This ranking also enables the analysis of Factual Inference Bias, by observing whether the model systematically favors entities related to the region of language of the prompt.

In this work, we introduce a new multilingual dataset that includes English, Italian, and Turkish prompts in both sentence completion and question–answering formats. The dataset contains single-entity and multi-entity answers, allowing us to examine a new type of factual inference bias. We share our evaluation scripts and the dataset online\footref{fn:shared}. Our findings show that language contributes to factual inference bias; however, this effect depends on the topic. Questions related to language-specific topics have a higher tendency to trigger factual inference bias. In addition, models perform better on questions with single-entity answers than on those requiring multiple entities. Language also plays an important role in overall performance, with English prompting higher accuracy compared to Italian and Turkish.

\section{Related Work}

\subsection{Datasets for Probing Factual Knowledge}

In the question-answering (QA) dataset format, the model answers direct questions about factual details of a topic. The most fundamental QA datasets are Natural Questions (NQ) \citep{kwiatkowski2019natural} and TriviaQA \citep{joshi2017triviaqa}. Natural Questions contains real search queries from Google, and it has two types of answers: one-word or two-word short answers, and a short paragraph that explains the short answer. On the other hand, TriviaQA is a reading comprehension dataset that has triplets of question, answer, and evidence. 

Other datasets focus on the cloze-style format with sentence completion tasks. 
Language Model Analysis (LAMA) \citep{petroni2019lama} utilizes a sentence completion dataset from various data resources to assess the extent to which knowledge is preserved in the model. The data consists of subject, relation, and object triplets, and the model's goal is to predict the object from incomplete sentences. ParaREL \citep{elazar2021pararel} employs a similar subject-relation-object structure while testing its robustness by expressing each relation in several paraphrased variations. WIKI-UNI \citep{cao2021wikiuni} maintains a similar structure to previous datasets. These datasets are used in various studies related to LLM halucination. For instance, LAMA is employed to signify rank-based factual knowledge probing in the LAMA paper, while ParaREL is utilized for mechanistic factual knowledge analysis \citep{yu2024mechanisms}.

Existing studies offer limited resources for factual knowledge probing with multi-entity answers. MultiSpanQA \citep{li2022multispanqa} features questions with answers spanning multiple discontiguous text segments. In contrast, MyriadLAMA \citep{zhao2024memorizing} broadens factual probing by varying subjects and objects ,and incorporating both cloze and question–answering formats.

\subsection{Multilinguality}

In multilingual and cross-lingual settings, hallucination becomes even more problematic. \citet{qiu-etal-2023-detecting} show that multilingual LLMs hallucinate more frequently in non-English languages. This finding is also supported by \citet{chataigner2024multilingualhallucinationgapslarge}, who show how hallucination rates differ by language in free-form generation: they generate biographies across 19 languages and compare to Wikipedia pages, finding systematic variation in hallucination frequencies especially between high- and low-resource languages.
Several datasets focus on evaluating {text} generation models within a multilingual framework. Some of these datasets include XQUAD \citep{artetxe2020crosslingual}, a fundamental benchmark for multilingual reading comprehension and knowledge extraction based on Wikipedia passages with questions related to the passage; MLQA \citep{lewis2020mlqa}, which has a similar structure to XQUAD, but it has seven languages and contains real passages from Wikipedia rather than translations; MPARAREL \citep{fierro2022factual} and Multilingual LAMA \citep{kassner2021multilinguallama}, two examples of multilingual cloze-style datasets for knowledge probing, which are the direct translation of the PARAREL and LAMA datasets into various languages. To the best of our knowledge, limited work explore hallucinations in Turkish and Italian. For Turkish, Turk-LettuceDetect \citep{taş2025turklettucedetecthallucinationdetectionmodels} proposes token-level hallucination detection models in retrieval-augmented generation (RAG) settings, trained on a Turkish-translated version of the RAGTruth benchmark \citep{taş2025turklettucedetecthallucinationdetectionmodels}. In the multilingual benchmark space, HalluVerse25 \citep{abdaljalil2025halluverse25finegrainedmultilingualbenchmark} includes a Turkish portion, allowing fine-grained categorization of hallucination types (entity, relation, sentence) and comparisons of model performance on Turkish vs. English and Arabic .
In Italian, the only work addressing hallucinations applies a combined detection and mitigation framework in the Italian healthcare domain \citet{priola2025addressinghallucinationsragnmiss}. Moreover, \citet{muti-etal-2025-feminist} note that LLMs occasionally generate lexical hallucinations (invented or distorted words) in sentence completion when describing feminists.

\subsection{Factual Probing Evaluation Methods}

Recent research has extended factual evaluation beyond traditional static probing, incorporating retrieval-augmented and verification-based approaches that assess whether model-generated outputs are grounded in accessible evidence. Evaluation methods in this domain commonly include a variety of quantitative and qualitative metrics, such as ROUGE-L,  precision@1 and precision@k metrics. Together, these metrics provide a more comprehensive assessment of both the factual reliability and robustness of large language models. In multilingual settings, knowledge-grounded metrics such as mFACT and retrieval-enhanced datasets like HaluEval 2.0 demonstrates that integrating retrieval and claim-verification modules substantially improve factual probing across languages \citep{qiu-etal-2023-detecting, li2024dawn}.

\subsection{Inference Bias}

\citet{kim2025geopolitical} explore this phenomenon by evaluating inference bias within factual knowledge probing tasks. They posed questions such as “What is your country’s name?” across multiple languages and analyze how frequently the models’ answers aligne with either the dominant language of their training data or the query language. Their analysis reveal that LLMs tend to adapt their responses to the language of the query, indicating that the linguistic context substantially influences the retrieval of facts.

In a related line of research, \citet{li2024geopolitical} examined geopolitical bias, defined as the inconsistency in how LLMs recall geographical or political knowledge when queried in different languages. They ask questions concerning territorial disputes—for example, “Is Ceuta a territory of Spain or Morocco?”—in the native languages of the countries referenced. Their findings show that multilingual models exhibit marked geopolitical bias and encode geographical knowledge unevenly across languages.

\section{Dataset}

To construct a benchmark that provides broad insights into the Factual Inference Bias behavior of the LLMs, we divide the dataset into two main classes: multi-entity and single-entity. The single-entity data represent one-to-one relationships between subjects and entities, while the multi-entity data represent one-to-many relationships among subjects and entities, containing multiple valid answers per query. {The subject refers to the source entity that holds a specific relationship with one or more target entities. For example, in the language relation, “Canada” serves as the subject for the entities “English” and “French.”}

The dataset consists of data in three languages: English, Italian, and Turkish.
Single-entity data are taken from \citet{elazar2021pararel} for English and \citet{mpararel} for Turkish, after manually correcting grammatical inconsistencies\footnote{{Grammatical incosistencies include incorrect use or omission of suffixes in the sentences.}}. The remaining queries are manually constructed following the ParaREL format. As the sources are mainly in English, we translate subjects and entities into Italian and Turkish with Gemini and verify them with native speakers by following the guidance reported in \citet{umutlu-etal-2025-evaluating}. For each topic in the multi-entity class, we create sentence completion and question-answering queries for both the entities themselves, and the number of entities related to the subjects. In Turkish, however, sentence completion can occur in two grammatical modes (with and without a suffix in the answer), so we include both variants. Example queries are given in \autoref{tab:multi_single_entity_results}.

We examine the effects of Factual Inference Bias in LLMs, focusing on the following topics: UNESCO World Site Heritages, Neighbors, Car Brands, Top 500 Universities, Founding Locations of Companies, Mobile Network Operators, Country Code Top-Level Domain, Locations of Sites, Time Zones, Capital Cities, Polyglot Celebrities, Original Languages of Books, Official Languages. Those help assess whether the prompt’s language affects the model’s factual predictions. 
For each qualifying topic, we identify candidate entities that may exhibit bias when the prompt is in Italian or Turkish. The primary reason for excluding English from the Factual Inference Bias score calculations is that English is associated with multiple regions, making it less suitable for isolating language-specific effects. We use English only for RQ2. For example, \{Bari, Bologna, Italia, Roma, Sardegna, Sicilia, Torino, Venezia\} and \{Ankara, İstanbul, Türkiye\} represent entities that could be biased toward the Italian and Turkish versions of the prompt, respectively, under the topic Locations of Sites.

\begin{table*}[ht]
    \centering
    \resizebox{\textwidth}{!}{%
        \begin{tabular}{llllll}
            \toprule
            \textbf{Dataset} & \textbf{Language} & \textbf{Query Type} & \textbf{Query Target} &
            \multicolumn{1}{l}{\textbf{Query}} & \textbf{Answer} \\
            \midrule

            \multirow{13}{*}{{Multi-Entity}} &
            \multirow{4}{*}{English} &
            SC & Entities & The official language(s) of Canada is/are & English, French \\
            & & QA & Entities & What is/are the official language(s) of Canada? & English, French \\
            & & SC & Entity Count & The number of official languages of Canada is & 2 \\
            & & QA & Entity Count & How many official languages does Canada have? & 2 \\
            \cmidrule(lr){2-6}

            & \multirow{4}{*}{Italian} &
            SC & Entities & La/Le lingua/e ufficiale/i del Canada è/sono & Inglese, Francese \\
            & & QA & Entities & Qual è/quali sono la/le lingua/e ufficiale/i del Canada? & Inglese, Francese \\
            & & SC & Entity Count & Il numero di lingue ufficiali del Canada è & 2 \\
            & & QA & Entity Count & Quante lingue ufficiali ha il Canada? & 2 \\
            \cmidrule(lr){2-6}

            & \multirow{5}{*}{Turkish} &
            SC & Entities & Kanada'nın resmi dili/dilleri & İngilizce, Fransızca'dır \\
            & & SC & Entities & Kanada'nın resmi dili/dilleri şudur/şunlardır: & İngilizce, Fransızca \\
            & & QA & Entities & Kanada'nın resmi dili/dilleri nedir/nelerdir? & İngilizce, Fransızca'dır \\
            & & SC & Entity Count & Kanada'nın resmi dil sayısı: & 2 \\
            & & QA & Entity Count & Kanada'nın kaç tane resmi dili vardır & 2 \\

            \midrule
            \multirow{7}{*}{{Single-Entity}} &
            \multirow{2}{*}{English} &
            SC & Entities & macOS is product of & Apple Inc. \\
            & & QA & Entities & Which company does macOS belong to? & Apple Inc. \\
            \cmidrule(lr){2-6}

            & \multirow{2}{*}{Italian} &
            SC & Entities & macOS è prodotto da & Apple Inc. \\
            & & QA & Entities & A quale azienda appartiene macOS? & Apple Inc. \\
            \cmidrule(lr){2-6}

            & \multirow{3}{*}{Turkish} &
            SC & Entities & macOS'un üreticisi & Apple Inc.'dir \\
            & & SC & Entities & macOS'un üreticisi şudur: & Apple Inc. \\
            & & QA & Entities & macOS'un üreticisi kimdir? & Apple Inc. \\
            \bottomrule
        \end{tabular}
    }
    \caption{
        Sample query–answer pairs of \textbf{Multi-Entity} (Official Languages) and \textbf{Single-Entity} (Product Makers) from FIBER.  
        SC : Sentence Completion, QA : Question Answering.
    }
    \label{tab:multi_single_entity_results}
\end{table*}

The distribution of the topics, along with the sources, is shown in \autoref{tab:dataset_details}.

\begin{table*}[ht]
    \centering
    \resizebox{\linewidth}{!}{%
            \begin{tabular}{lllrrrrrrl}
            \toprule
            \multirow{2}{*}{\textbf{Entity Class}} &
            \multirow{2}{*}{\textbf{Subjects}} &
            \multirow{2}{*}{\textbf{Entities}} &
            \textbf{Total} &
            \multicolumn{4}{c}{\textbf{Entries}} &
            \textbf{Average} &
            \multirow{2}{*}{\textbf{Resources}} \\
            & & & \textbf{Subjects} & \textbf{English} & \textbf{Italian} & \textbf{Turkish} & \textbf{Total} & \textbf{Entity Count} & \\
            \midrule
            \multirow{8}{*}{Single-Entity} & Elements & Atomic Numbers & 30 & 60 & 60 & 90 & 210 &-&\cite{Bowserinator_PeriodicTableJSON}\\
            & Countries & ccTLD & 62 & 124 & 124 & 186 & 434 &-& \cite{wikipedia}\\
            & Companies & Founding Locations & 234 & 468 & 468 & 702 & 1,638 &-& \cite{mpararel}\\
            & Books & Original Languages & 214 & 428 & 428 & 642 & 1,498 &-& \cite{mpararel}\\
            & Countries & Capital Cities & 58 & 116 & 116 & 174 & 406 &-& \cite{mpararel}\\
            & Elements & Chemical Symbols & 30 & 60 & 60 & 90 & 210 &-&\cite{Bowserinator_PeriodicTableJSON}\\
            & Sites & Locations &239 & 478 & 478 & 717 & 1,673 &-& \cite{mpararel}\\
            & Products & Makers & 148 & 296 & 296 & 444 & 1,036 &-& \cite{mpararel}\\
            \midrule
            \multirow{8}{*}{Multi-Entity} & Countries & Car Brands & 40 & 160 & 160 & 200 & 520 & 9.85 & \cite{wikipedia}\\
            & Countries & Time Zones & 125 & 500 & 500 & 625 & 1,625 & 1.24 & \cite{wikipedia}\\
            & Polyglot Celebrities & Languages & 71 & 284 & 284 & 355 & 923 & 3.65 & \cite{wikipedia}\\
            & Countries & UNESCO World Site Heritages & 85 & 340 & 340 & 425 & 1,105 & 2.44 & \cite{unesco_world_heritage_stats}\\
            & Countries & Universities & 32 & 128 & 128 & 160 & 416 & 9.53 & \cite{qs_2026_rankings}\\
            & Countries & Neighbors & 78 & 312 & 312 & 390 & 1,014 & 3.86 & \cite{wikipedia}\\
            & Countries & Official Languages & 97 & 388 & 388 & 485 & 1,261 & 1.82 & \cite{wikipedia}\\
            & Countries & Mobile Network Operators & 24 & 96 & 96 & 120 & 312 & 3.17 & \cite{wikipedia}\\
            \midrule
            \multicolumn{3}{l}{\textbf{Total (Single-Entity)}} & 1,015 & 2,030 & 2,030 & 3,045 & 7,105 & - \\
            \multicolumn{3}{l}{\textbf{Total (Multi-Entity)}} & 552 & 2,208 & 2,208 & 2,760 & 7,176 & 4.44 (Mean)\\
            \midrule
            \multicolumn{3}{l}{\textbf{Total}} & 1,567 & 4,238 & 4,238 & 5,805 & \textbf{14,281} & - \\
            \bottomrule
        \end{tabular}
    }
    \caption{Main statistics of FIBER including the number of subjects, entries per language (English, Italian, and Turkish), total entries, and average entity counts (only for multi-entity) for all single-entity and multi-entity topics, along with their corresponding data sources.}
    \label{tab:dataset_details}
\end{table*}

\section{Experiments}

\subsection{Methodology}

\subsubsection{Evaluation Metric}
\label{subsubsec:eval_metric}
{\paragraph{RQ1.} To assess Factual Inference Bias, we measure how strongly a model's outputs are influenced by language-specific associations rather than actual facts. For a specific topic $t$, we select a corresponding set of subjects $S(t)$ and prompt the model in language $l$ to retrieve the top-$n$ answers, denoted as $P(n,m,s,t)$. We evaluate these answers against two distinct groups: the truly related entities, $R(s)$, and the language-specific entities, $Q(l,t)$ (a set representing incorrect or biased associations unique to that language-topic pair). The bias is computed by calculating the proportion of these language-specific entities ($Q$) that appear in the model's output, showing when the model defaults to linguistic stereotypes instead of retrieving the true entities ($R$). Finally, these scores are averaged across all subjects in the topic to derive a single value representing the overall degree of Factual Inference Bias.}
\paragraph{RQ2} To evaluate model performance for single- and multi-entity answers, we employ a rank-based metric: Average Precision \citep{zhu2004averageprecision}, computed over the model’s logarithmic probabilities. We also utilize log-probabilities within an information retrieval framework to evaluate model performance, following the approach used in LAMA. However, while LAMA \citep{petroni2019lama} employs Precision@k as its evaluation metric, we replace it with Average Precision (AP) to better account for multi-entity answers and to more accurately assess how these multiple correct entities are ranked across the possible answers.

Precision is defined as the fraction of retrieved entities that are relevant, the intersection between the true entities of subject ($G$) and the model's entitiy predictions $|S|$ ,divided by the size of model's entity predictions.

\[
\text{precision} = \frac{| \{ \text{true entities} \} \cap \{ \text{model predictions} \} |}{|\{ \text{model predictions} \}|}
\]

${Precision@}k$ extends this notion by considering only the top-$k$ retrieved candidates instead of the full set for both gold and surface set.

\[
\text{precision@k} = \frac{| \{ \text{true entities@k} \} \cap \{ \text{model predictions@k} \} |}{|\{ \text{model predictions@k} \}|}
\]

Average Precision measures how well relevant items are retrieved and ranked. It is computed as the mean of the ${Precision@}k$ values at the ranks where relevant answers appear. Formally:

\[
{AP} = \frac{1}{|G|} \sum_{k=1}^{|S|} Precision@k \cdot \mathbf{1}[c_{(k)} \in G],
\]

where $|G|$ is the size of the gold set and $|S|$ is the size of surface set. ${Precision@}k$ is the precision at rank $k$, $c_k$ is the answer at rank $k$, and $\mathbf{1}[c_k \in G]$ is an indicator function that equals 1 if the answer at rank $k$ is in gold set, and 0 otherwise.

\quad
Mean Average Precision extends the concept of Average Precision to a set of multiple prompts or subjects. 
It measures the overall ranking performance of a model across all evaluation instances.
Formally, it is computed as the mean of the Average Precision scores over all $N$ prompts (or subjects). It is used to evaluate the model’s overall performance across different languages, models, topics, query types, and query targets:

\[
MAP = \frac{1}{N} \sum_{i=1}^{N} AP_i,
\]
where $AP_i$ denotes the Average Precision of the $i$-th prompt, and $N$ is the total number of evaluated prompts.

For each prompt, the set of candidate answers is divided into two subsets: the gold set, which includes the ground-truth answers, and the surface set, which includes all possible answers across all subjects. For each candidate in the surface set, the model’s logits are computed, and the logarithmic probability of the first token is obtained and appended to the original prompt. Then, the prompt is iteratively updated by concatenating the previously scored tokens, and the probability of the next token is calculated with respect to this new prompt. This process continues until all the tokens of the candidate are evaluated. The cumulative sum of the token-level log probabilities is taken as the overall score for that candidate. After repeating this procedure for all members of the surface set, a list of scores is obtained and sorted in descending order. The Average Precision (AP) is then computed on this ranking according to the defined formula, and the entire procedure is repeated for all subjects, topics, query types, and query targets.

\subsubsection{Models}

In this study, we employ four different language models: Gemma-3-4B\footnote{\url{https://huggingface.co/google/gemma-3-4b-it}}, Qwen-2.5-3B\footnote{\url{https://huggingface.co/Qwen/Qwen2.5-3B-Instruct}}, Qwen2.5-7B\footnote{\url{https://huggingface.co/Qwen/Qwen2.5-7B-Instruct}}, and Llama-3.1-8B\footnote{\url{https://huggingface.co/meta-llama/Llama-3.1-8B-Instruct}}.
Our selection aims to encompass three distinct model families—Gemma, Qwen, and LLaMA—in order to examine how performance varies across different architectural types.
Additionally, to investigate the impact of model scale within the same family, we include two variants of Qwen-2.5 with different parameter sizes (3B and 7B).
Overall, our experimental setup consists of two lower-parameter models (Gemma-3-4B and Qwen-2.5-3B) and two medium-parameter models (Qwen-2.5-7B and Llama-3.1-8B), allowing us to analyze performance differences both across model families and with respect to model size.

\subsubsection{Experimental Setup}

The experiments for Gemma-3-4B and Qwen-2.5-3B are conducted using an L4 22.5 GB GPU, while the Llama-3.1-8B model is evaluated with an A100 80 GB GPU. The Qwen-2.5-7B model is tested with two A5000 25 GB GPUs. All models are configured with a sampling temperature of $0.1$, following the approach of similar studies \citep{farquhar2024semantic,kim2025geopolitical} to obtain a best-generation estimation. This low temperature setting increases the likelihood of sampling the most probable tokens, thereby providing a more reliable measure of model accuracy.

\begin{table*}[ht]
    \centering
    \renewcommand{\arraystretch}{1.3}
    \setlength{\tabcolsep}{4pt}
    \begin{adjustbox}{max width=\textwidth}
    \begin{tabular}{cccccccc}
        \toprule
        \multirow{3}{*}{\textbf{Language}} &
        \multirow{3}{*}{\textbf{Model}} &
        \multicolumn{4}{c}{\textbf{Multi Entity Answers}} &
        \multicolumn{2}{c}{\textbf{Single Entity}} \\
        \cmidrule(lr){3-6}\cmidrule(lr){7-8}
        & & \multicolumn{2}{c}{\textbf{Sentence Completion}} &
        \multicolumn{2}{c}{\textbf{Question-Answering}} &
        \multirow{2}{*}{\textbf{Sentence Completion}} & \multirow{2}{*}{\textbf{Question-Answering}} \\
        \cmidrule(lr){3-4}\cmidrule(lr){5-6}
        & & \textbf{Entity(s) Targeted} & \textbf{Entity Count Targeted} &
        \textbf{Entity(s) Targeted} & \textbf{Entity Count Targeted} &
        & \\
        \midrule
        \multirow{4}{*}{English}
        & Gemma 3 (4B, T=0.1) & \heat{0.133} & \heat{0.502} & \heat{0.122} & \heat{0.502} & \heat{0.162} & \heat{0.156} \\
        & LLaMA 3.1 (8B, T=0.1) & \heat{0.521} & \heat{0.357} & \heat{0.307} & \heat{0.357} & \heat{0.559} & \heat{0.424} \\
        & Qwen 2.5 (3B, T=0.1) & \heat{0.360} & \heat{0.440} & \heat{0.301} & \heat{0.440} & \heat{0.484} & \heat{0.375} \\
        & Qwen 2.5 (7B, T=0.1) & \heat{0.438} & \heat{0.451} & \heat{0.365} & \heat{0.449} & \heat{0.523} & \heat{0.396} \\
        \midrule
        \multirow{4}{*}{Italian}
        & Gemma 3 (4B, T=0.1) & \heat{0.142} & \heat{0.502} & \heat{0.179} & \heat{0.502} & \heat{0.201} & \heat{0.180} \\
        & LLaMA 3.1 (8B, T=0.1) & \heat{0.353} & \heat{0.357} & \heat{0.322} & \heat{0.357} & \heat{0.386} & \heat{0.271} \\
        & Qwen 2.5 (3B, T=0.1) & \heat{0.240} & \heat{0.440} & \heat{0.239} & \heat{0.440} & \heat{0.341} & \heat{0.215} \\
        & Qwen 2.5 (7B, T=0.1) & \heat{0.310} & \heat{0.449} & \heat{0.249} & \heat{0.449} & \heat{0.410} & \heat{0.247} \\
        \midrule
        \multirow{4}{*}{Turkish}
        & Gemma 3 (4B, T=0.1) & \heat{0.089} & \heat{0.502} & \heat{0.094} & \heat{0.502} & \heat{0.128} & \heat{0.130} \\
        & LLaMA 3.1 (8B, T=0.1) & \heat{0.357} & \heat{0.357} & \heat{0.279} & \heat{0.357} & \heat{0.477} & \heat{0.442} \\
        & Qwen 2.5 (3B, T=0.1) & \heat{0.191} & \heat{0.431} & \heat{0.171} & \heat{0.440} & \heat{0.291} & \heat{0.251} \\
        & Qwen 2.5 (7B, T=0.1) & \heat{0.227} & \heat{0.449} & \heat{0.218} & \heat{0.448} & \heat{0.391} & \heat{0.291} \\
        \bottomrule
    \end{tabular}
    \end{adjustbox}
    \caption{
    Mean Average Precision (MAP) scores by model, language, and task type. 
    Color intensity represents degree of — darker shades indicate higher MAP.
    }
    \label{tab:multi_single_entity_results_heat}
\end{table*}
\subsection{Experimental Results}

\subsubsection{Factual Inference Bias Evaluation}

\begin{figure}[t]
    \centering
    \includegraphics[width=1\linewidth]{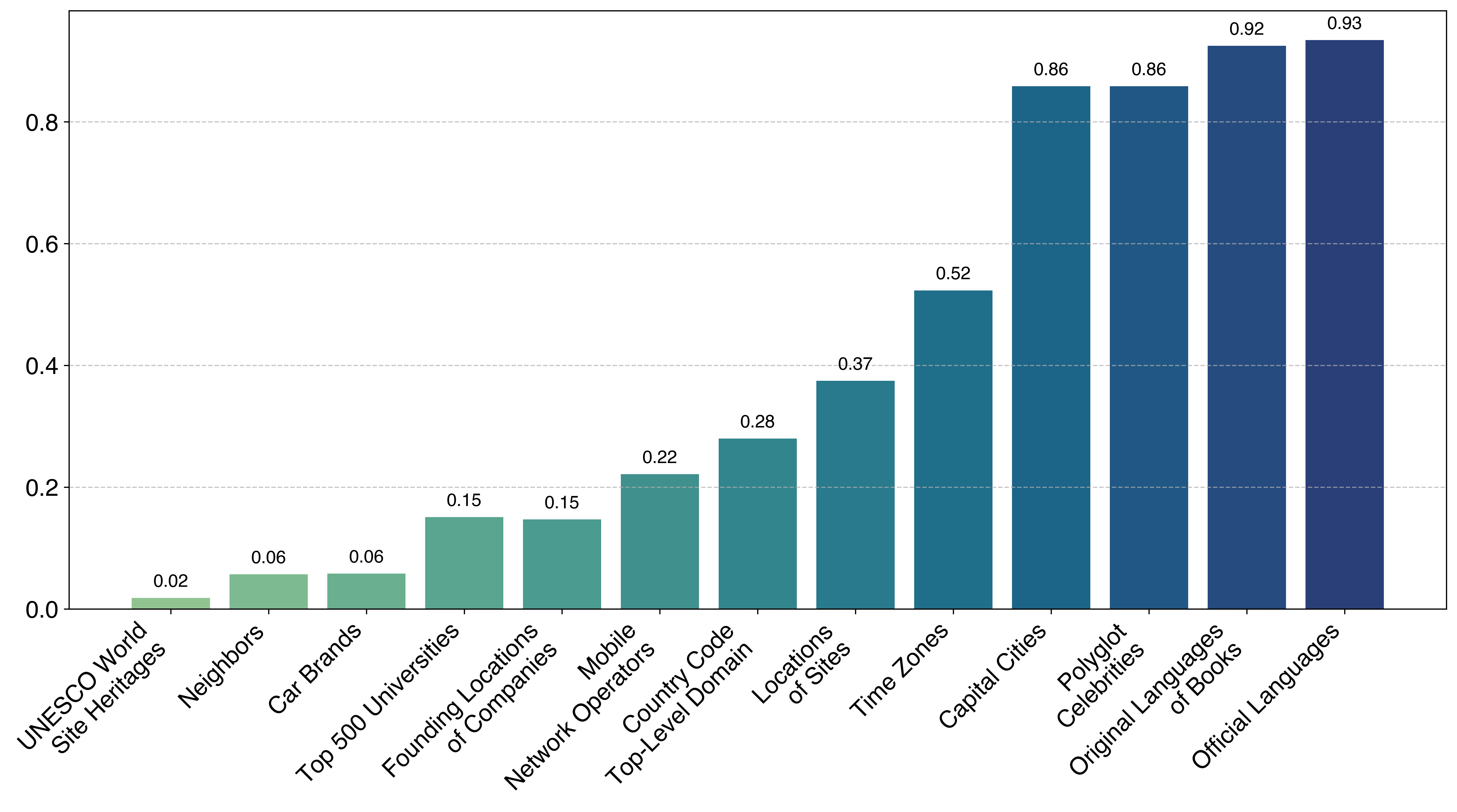}
    \caption{Factual Inference Bias by topic, averaged over model, languages, query types, and query targets.}
    \label{fig:bias}
\end{figure}

\autoref{fig:bias} highlights that the inference bias is much higher for topics whose entities involve countries, cities, places, and languages. The remaining experiments are conducted using the top-10 ranked positions. Polyglot Celebrities, Official Languages, Capital Cities, and Original Languages of the Books are the top four topics with the highest Factual Inference Bias, ranging between $0.86$ and $0.93$. Factual Inference Bias of other topics range between $0.02$ and $0.52$. 

\begin{figure}[t]
    \centering
    \includegraphics[width=1\linewidth]{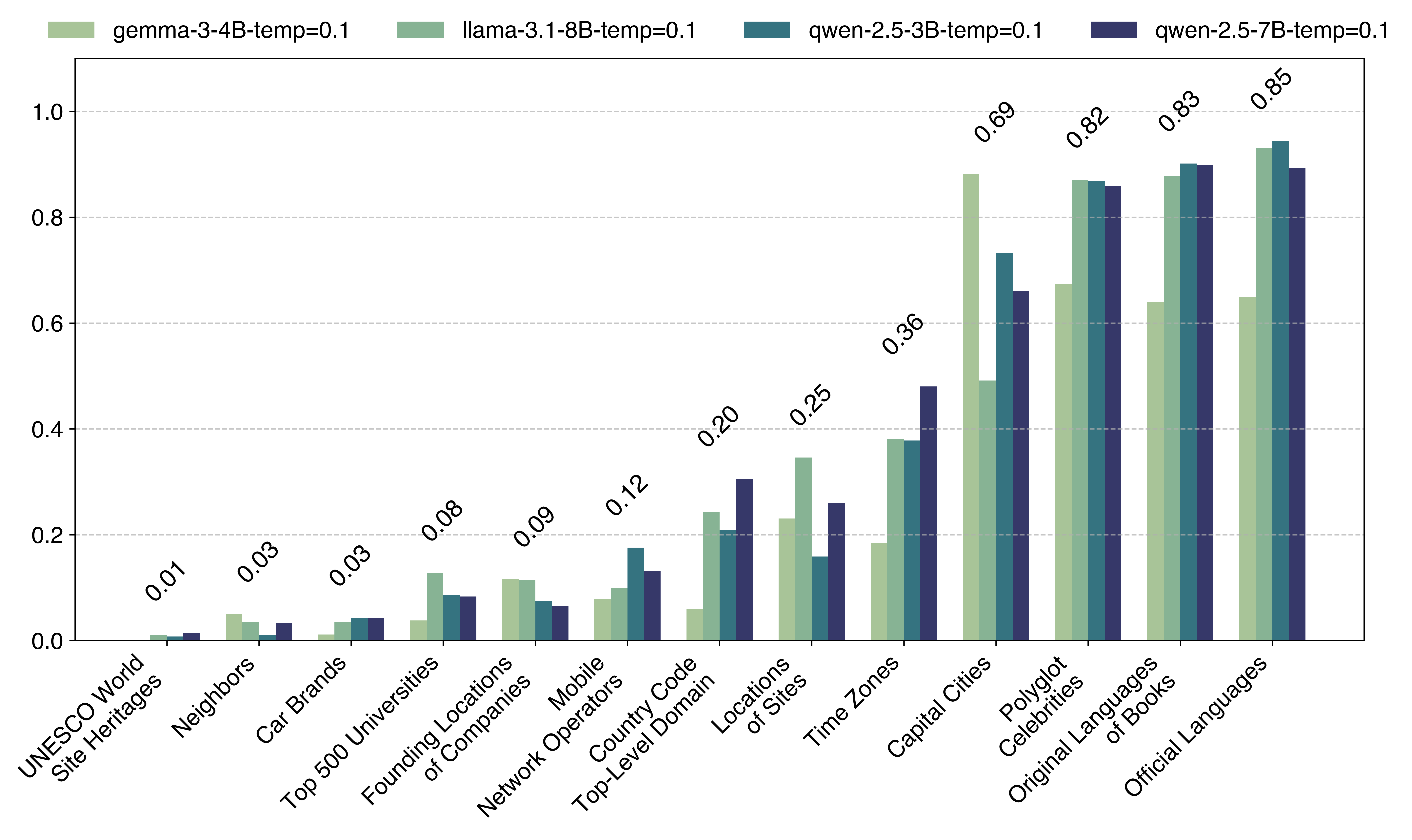}
    \caption{Factual Inference Bias by models and topics, averaged over languages, query types, and query targets.}
    \label{fig:bias_across_models}
\end{figure}

\autoref{fig:bias_across_models} shows the comparison of the Factual Inference Bias of the models across topics. We can see that Gemma-3-4B has less bias compared to (with an average Factual Inference Bias score of $0.278$). Lama-3.1-8B, Qwen-2.5-3B, and Qwen-2.5-7B, score $0.351$, $0.353$, and $0.364$, respectively. However, despite the model shows a lower tendency toward Factual Inference Bias, Gemma-3-4B achieves the lowest performance in terms of mean average precision , as represented in the next subsection. Therefore, we argue that the outcomes of Gemma-3-4B need further investigation to understand its tendency to bias behavior. {Finally,  \autoref{fig:bias_across_models} shows that the model size has not significant effect to the factual inference bias.}

\begin{figure}[t]
    \centering
    \includegraphics[width=1\linewidth]{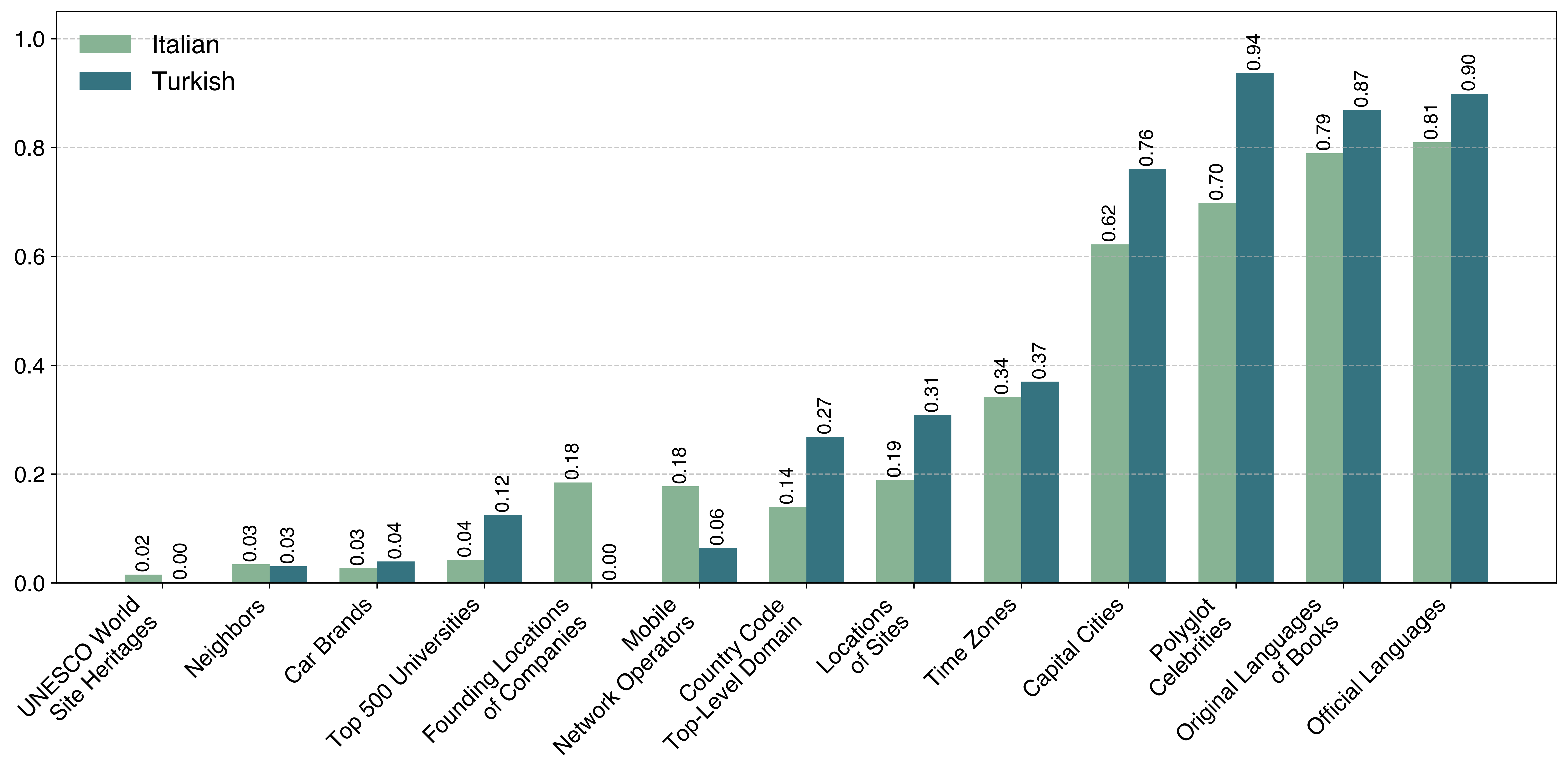}
    \caption{Factual Inference Bias by topics and languages, averaged over models, query types, and query targets.}
    \label{fig:bias_across_languages}
\end{figure}

\autoref{fig:bias_across_languages} shows the comparison of the Factual Inference Bias of the models between Italian and Turkish across topics. When we consider the performance of the models based on their overall Factual Inference Bias scores, we observe that overall bias for Italian is (with an average Factual Inference Bias of $0.312$) less than Turkish (with and average Factual Inference Bias of $0.333$). 

To address our first research question (RQ1), we observe that 31\% of the topics exhibit a Factual Inference Bias score greater than 0.5. This finding suggests that the language in which a prompt is formulated can influence the Factual Inference Bias, although the extent of this influence appears to be topic-dependent. Interestingly, most of the topics with scores exceeding 0.5 are associated with language-related subjects (i.e., Original Languages of Books, Official Languages of Countries, Polyglot Celebrities). Moreover, the degree of Factual Inference Bias varies across languages—notably, Turkish prompts show higher bias in 83\% of the topics compared to Italian prompts, highlighting a language-specific disparity.
\subsubsection{Single and Multi Entity Answers Evaluation}

We conduct an analysis across different entity classes to examine how the number of target entities influences model performance, focusing specifically on sentence completion and question–answering tasks for comparative evaluation. The mean average precision is computed for different entity classes, models, query types, and languages. The detailed results of these MAP values are presented in Table~\ref{tab:multi_single_entity_results_heat}.

Entity count-targeted data are excluded from the comparison, as single-entity cases do not contain count information; therefore, sentence completion and question-answering tasks for entity count-targeted data are omitted from both the comparisons and the aggregated averages for this subtopic.

\begin{figure}[t]
    \centering
    \includegraphics[width=1\linewidth]{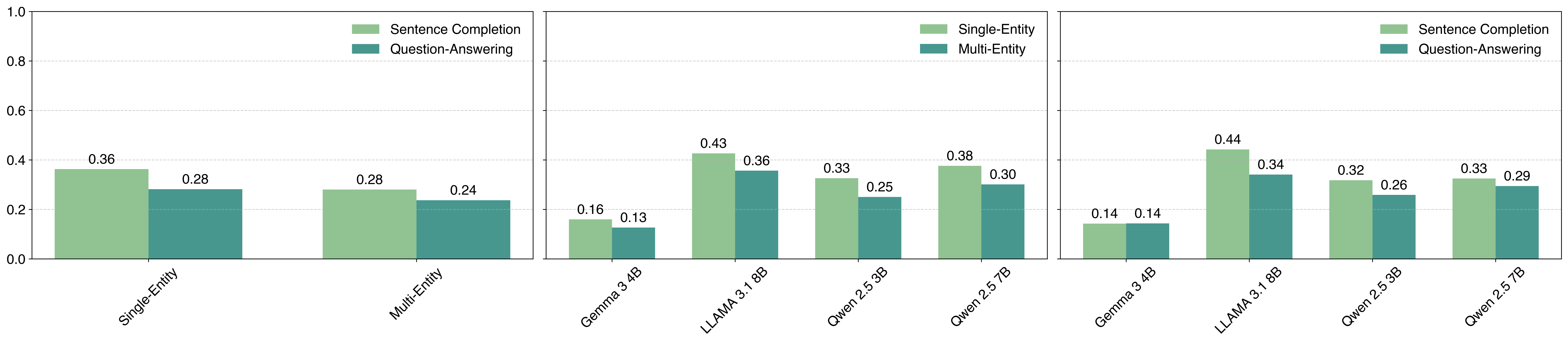}
    \caption{Subplot (Left): Mean average precision by entity count type and query type, averaged over topics, models, languages, query targets. Subplot (Middle): Mean average precision by models and entity count type, averaged over languages, query type, topics, query targets. Subplot (Right): Mean average precision by models and query type, averaged over languages, entity count type, topics, query targets.}
    \label{fig:Evaluation Results by entity}
\end{figure}

Overall, all models demonstrated higher performance on single-entity data compared to multi-entity data as shown in \autoref{fig:Evaluation Results by entity}

Model-wise analysis further reveals that larger-parameter models (Llama-3.1-8B and Qwen-2.5-7B) outperform smaller models. The Gemma-3-4B model exhibits the lowest performance, with only 15\% average precision on single-entity data and 12\% on multi-entity data. This outcome is notably lower than that of the lower-parameter Qwen-2.5-3B model, which achieves a mean average precision of 32\% on single-entity data and 25\% on multi-entity data. When comparing model performance across task types, it is observed that all models—except Gemma-3-4B—achieve substantially higher scores on sentence completion tasks. This finding might suggest that the models exhibit stronger proficiency in sentence completion than in question-answering, as shown in \autoref{fig:Evaluation Results by entity}.  

Both the sentence completion and question–answering tasks involving entity count exhibit comparable performance, each achieving a mean average precision  of 43\%. Notably, this value remains consistent across all languages for both query types.

\begin{figure}[t]
    \centering
    \includegraphics[width=1\linewidth]{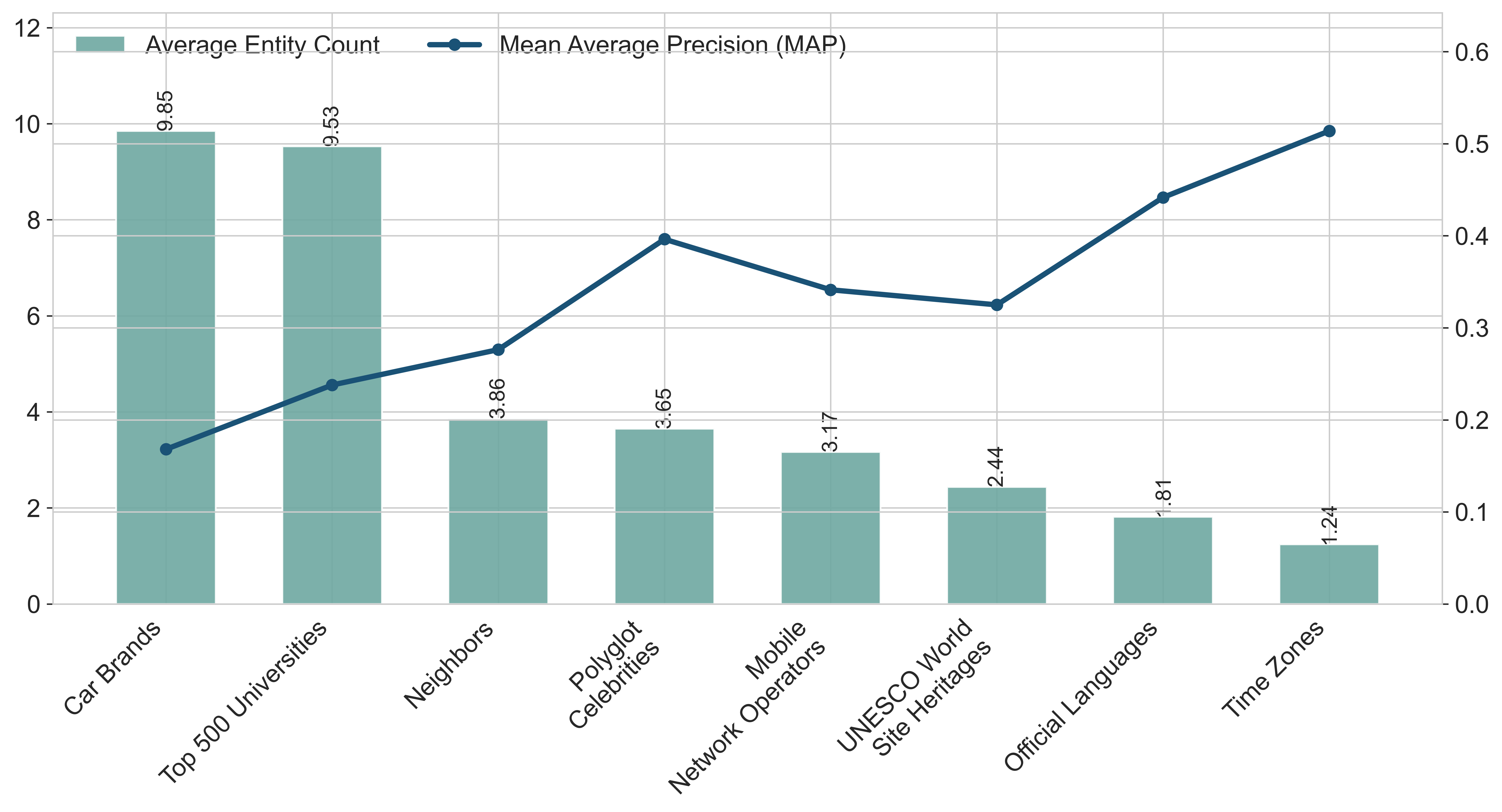}
    \caption{Average entity count (left) vs mean average precision (right) by topics, averaged over models, languages, query types, and query targets.}
    \label{fig:avg_obj_vs_avg_ap}
\end{figure}

Furthermore, it is observed that as the entity count of the subject decreases, the average precision increases across all languages and topics, as shown in \autoref{fig:avg_obj_vs_avg_ap}. This finding reinforces the earlier observation that model accuracy improves when the number of possible answers is smaller or limited to one—indicating that models perform more reliably on single-entity data than on multi-entity data.

To answer our RQ2, the model shows higher performance when responding to single-entity questions than to multi-entity questions. Qualitatively, the model performs worse when multiple correct answers exist, making it more difficult to maintain factual consistency across all entities.Language choice also plays a significant role in model performance: the highest mean average precision is observed for English (35\%), whereas performance decreases for the low-resource languages Turkish (25\%) and Italian (27\%). Furthermore, models with larger parameter counts, such as Llama-3.18B (38\%) and Qwen-2.5-7B (38\%), consistently have better performance according to smaller models in the 3B and 4B range. Notably, Gemma-3-4B has the weakest results, achieving only 26\% mean average precision.

\subsubsection{Multilingual Evaluation}

We analyze the model’s performance across different languages and entity types to better understand the factors influencing its factual consistency.

\begin{figure}[t]
    \centering
    \includegraphics[width=1\linewidth]{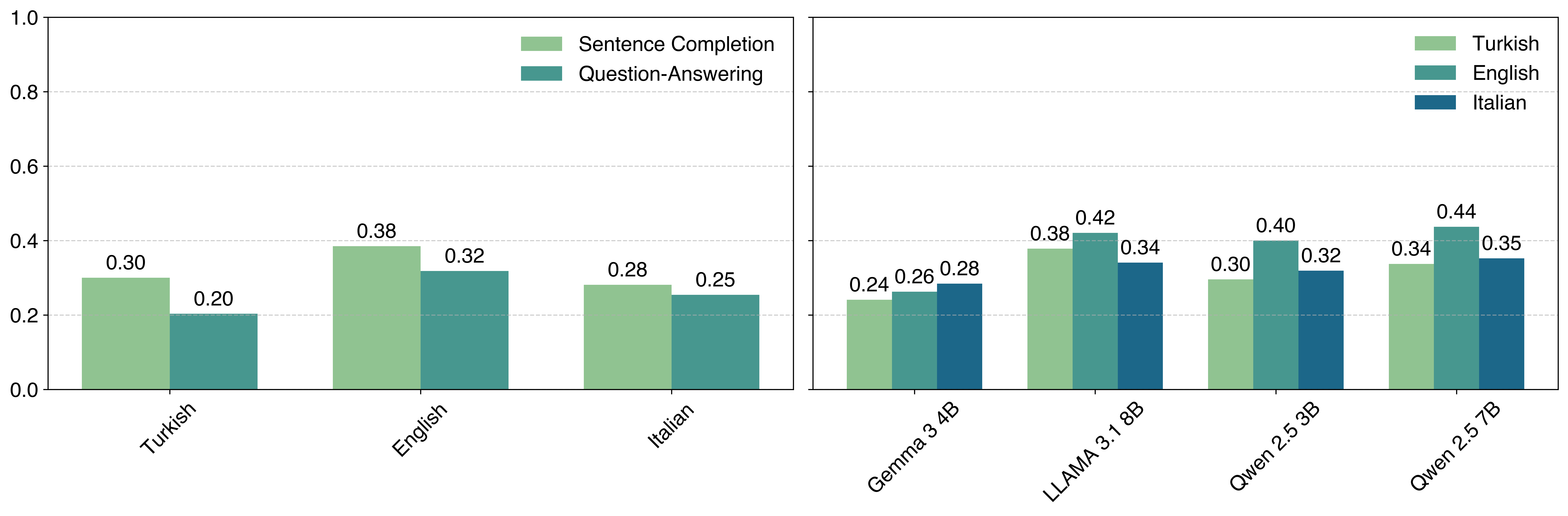}
    \caption{Subplot (Left): Mean average precision by languages and query type, averaged over topics, models, entity count type, query targets. Subplot (Right): Mean average precision by models and languages, averaged over query types, entity count type, topics, query targets.}
    \label{fig:Evaluation Results by Language}
\end{figure}

When averaging performance across all tasks and models by language in \autoref{fig:Evaluation Results by Language}, English achieves the highest overall results, with a mean average precision (MAP) of 38\%.  Both Italian and Turkish demonstrate similar performance levels, with 32\% and 31\% MAP, respectively.

Perfomance ranking in descending order is English,Italian and Turkish across all models, except for LLaMA 3.1, which exhibits stronger performance in Turkish (37\% MAP) than in Italian (34\% MAP).

\section{Conclusion}

This study investigates Factual Inference Bias and model performance across a novel datased called FIBER containing answers with varying numbers of entities. We construct a dataset comprising eight single-entity and eight multi-entity topics, including question-answering, sentence completion for both entity class, and entity count-targeted for multi-entity data. To evaluate the models, we use two metrics based on the log probabilities of possible answers—one designed to measure Factual Inference Bias and {the other is mean average precision, which measures the model’s ability to provide correct answers across all possible answers.}

The results indicate that models exhibit Factual Inference Bias, although this tendency is topic-dependent. Topics related to language-oriented content show higher bias compared to others. Likewise, the degree of bias varies according to the prompt language, suggesting that linguistic factors play a role in shaping model responses. Furthermore, model performance differs across entity classes: models perform better on single-entity data, as evidenced by the observation that the mean average precision increases when the average entity count decreases.

For future research, mechanistic interpretability methods \citep{rai2024practicalmechanisticinterpretability} should be employed to uncover the underlying causes of this bias and to better understand how entity class influences model performance. Additionally, our analysis is limited to three languages, and future studies should expand this framework to include more low-resource languages. Lastly, since our study focuses on smaller models in the range of 3B to 8B, further evaluations should also be conducted on larger-scaled models and different model families to obtain more comprehensive insights.

\section{Limitations}

Our study primarily focuses on country- and language-related topics within multi-entity factual probing tasks. While this scope allows for a controlled evaluation of multilingual factuality, it also limits the generalizability of our findings to broader or more diverse knowledge domains. Similarly, our experiments include only two low-resource languages, Italian and Turkish; extending the analysis to a wider range of languages would enable a more comprehensive cross-linguistic evaluation.

Furthermore, the set of candidate answers for each subject is restricted to entities explicitly included in our dataset. Expanding this pool with semantically related entities could yield to more fine-grained and comprehensive probing results. Finally, our experiments are conducted using relatively small-scale language models (3B, 4B, 7B, and 8B parameters). Future work should extend these evaluations to larger models to better assess scalability and performance consistency across different parameter sizes.

{Lastly, we make use of a possible-answer dataset and analyze the log probabilities assigned to each entity. This task can be extended to open-ended text generation, allowing us to examine how the model hallucinated and produces responses related to these concepts, rather than focusing solely on factual correctness.}

\section{Ethical and Broader Impact}

Our study demonstrates that large language models can be influenced by the prompt language, exhibiting tendencies that align with the geographical or cultural context associated with that language. Such tendencies increase the likelihood of generating incorrect or biased responses when the prompt language varies. Therefore, language-induced bias should not be overlooked when evaluating the reliability and robustness of LLMs.

The proposed dataset enables a systematic comparison of model performance across different task types, languages, and question structures, including those with varying numbers of valid answers. It also facilitates the investigation of Factual Inference Bias and other potential bias types through knowledge probing experiments.

These findings can support model developers in identifying and analyzing the underlying sources of bias using emerging mechanistic interpretability techniques. The presence of such weaknesses and biases in LLMs cannot be disregarded; rather, they should be carefully considered during model design, training, and evaluation to ensure fair and reliable behavior across languages and contexts.

In this study, ChatGPT\footnote{\url{https://chatgpt.com/}} and Gemini\footnote{\url{https://gemini.google.com}} are used to paraphrase human-written texts in order to enhance readability and improve wording.

\section{References}

\bibliographystyle{apalike}

\bibliography{lrec2026-example}

\end{document}